\documentclass{article}

\usepackage[preprint]{neurips_2025}

\usepackage[utf8]{inputenc} 
\usepackage[T1]{fontenc}    
\usepackage{hyperref}       
\usepackage{url}            
\usepackage{booktabs}       
\usepackage{amsfonts}       
\usepackage{nicefrac}       
\usepackage{microtype}      
\usepackage{xcolor}         
\usepackage[pdftex]{graphicx}
\usepackage{subfigure}
\usepackage{bm}

\title{Spiking Neural Networks with Random Network Architecture}

\author{%
  Zihan Dai \\
  School of mathematical and science\\
  Soochow University\\
  Suzhou City, Jiangsu Province, PR China, 215006 \\
  \texttt{20234207017@stu.suda.edu.cn} \\
  Huanfei Ma \\
  School of mathematical and science\\
  Soochow University\\
  Suzhou City, Jiangsu Province, PR China, 215006 \\
  \texttt{hfma@suda.edu.cn} \\
}

\begin{document}

\maketitle

\begin{abstract}
  The spiking neural network, known as the third generation neural network, is an important network paradigm. Due to its mode of information propagation that follows biological rationality, the spiking neural network has strong energy efficiency and has advantages in complex high-energy application scenarios. However, unlike the artificial neural network (ANN) which has a mature and unified framework, the SNN models and training methods have not yet been widely unified due to the discontinuous and non-differentiable property of the firing mechanism. Although several algorithms for training spiking neural networks have been proposed in the subsequent development process, some fundamental issues remain unsolved. Inspired by random network design, this work proposes a new architecture for spiking neural networks, RanSNN,  where only part of the network weights need training and all the classic training methods can be adopted. Compared with traditional training methods for spiking neural networks, it greatly improves the training efficiency while ensuring the training performance, and also has good versatility and stability as validated by benchmark tests. 
\end{abstract}

\section{Introduction}
Spiking Neural Networks(SNNs) are constructed based on the firing mechanism of biological neurons \cite{MAASS19971659}. Unlike other types of artificial neural networks, the neurons in an SNN do not simply perform weighted summation of inputs and output through an activation function, instead, within an SNN the information is transmitted in the form of spike trains. Such a time-based encoding method is more consistent with the working principle of the biological nervous system compared to the continuous value encoding of traditional neural networks \cite{yamazaki2022spiking}. Therefore, taking inspiration from the biological system makes SNN a highly energy-efficient solution to the problem of signal processing.

Despite the remarkable progress in recent years \cite{taherkhani2020review}, the training algorithm of SNNs remains an open challenge. In order to realize the information process job, neural networks are typically trained using an optimization procedure, such as gradient descent algorithm, in which the parameters or weights are adjusted to minimize an objective function. When it comes to SNNs, the dynamics of hidden neurons induced by the firing mechanism is nondifferentiable or even discontinuous, preventing all the gradient-based optimization for training. To overcome this problem concerned with discontinuous spiking nonlinearity, several approaches have been studied with varying degrees of success. The first category  resort to biologically inspired local learning rules for the hidden neurons, such as Spike-Timing Dependent Plasticity (STDP) learning algorithm\cite{TAVANAEI201947}\cite{kheradpisheh2018stdp}\cite{diehl2015unsupervised}. Such local learning rules, which is mainly regulated by the local spike timing of the presynaptic and postsynaptic neurons, lacks the ability to process global information. Meanwhile, this learning method, which is difficult to be applied to complex tasks, suffers from the problem of unstable algorithm convergence. The second category seeks smoothing the network so that the backpropagation algorithm can works. To this, some efforts have been made to replace the commonly adopted leaky integrated-and-fire(LIF) model with  smooth spiking generating process such as Hodgkin-Huxley (HH), or FitzHugh-Nagumo(FHN) models. However, due to the high complexity, only few applications have been  reported with success using continuous-valued gating functions\cite{giannari2022model}. Alternatively, surrogate gradient(SG) method introduces backpropagation through time (BPTT) for training recurrent neural networks (RNNs) into SNNs using the gradient of a differentiable surrogate function and making it possible to directly train a large scale of SNN efficiently with LIF models\cite{wang2024brain}. However, the SG strategy makes the gradient prone to exploding or vanishing, leading to network degradation\cite{8891809}. Therefore, how to train the SNN effectively and accurately is still a open challenge.

On the other hand, in the recent years, a novel class of lightweight network architectures based on random generation mechanisms has emerged with distinctive advantages . Representative solutions, like reservoir computing (RC) and extreme learning machines (ELMs), employ random initialization and fixation of partial weight parameters, transforming traditional back-propagation-based training processes into linear system solving. Moreover, the random feature method has also attracted great attentions for solving PDE\cite{chen2022bridgingtraditionalmachinelearningbased}. These approaches preserve model representational capacity while significantly simplifying training procedures and  establish innovative frameworks for efficient neural network training in typical applications including time series analysis \cite{shao2023data}\cite{duan2023embedding}\cite{butcher2013reservoir}, speech recognition\cite{gallicchio2018design}\cite{skowronski2007automatic}\cite{liu2018speech}, image classification\cite{tong2018reservoir}\cite{park2019convolutional}\cite{cao2016extreme}, bio-informatics processing\cite{wang2008protein}\cite{wang2014fast}\cite{lacy2018using}, and so on.

Inspired by the architecture of random networks, in this work we propose a new framework of spiking neural networks. Within this framework, the hidden layers are randomly generated and fixed with certain distribution and thus there is no need to calculate the gradient in the hidden layers to propagate the loss. Validated by several benchmark tasks, we show that compared with traditional training methods for spiking neural networks, it greatly improves the training efficiency and also ensures good network performance.

\section{Results}
\subsection{Random Spiking Neural Networks}\label{sec:method}
In this section, the random architecture for SNN will be introduced. Specifically, we will consider a representative SNN framework with Leaky Integrate-and-Fire(LIF) neurons. The LIF model is commonly used as the network neuron unit to simulate the behavior of information propagation in the network, and the dynamics of one LIF neuron could be described by the following equations:
\begin{equation}\label{LIF}
    \left\{
    \begin{array}{rl}
    \displaystyle \tau \frac{\mathrm{d}{u}(t)}{\mathrm{d}t} &=- {u}(t) + rI(t) \\
    {s}(t) &= 1,{\rm if}\:{u}(t)>{u}_{thr}\\
    {s}(t) &= 0,{\rm otherwise}\\
    \end{array}
    \right.
\end{equation}
where ${u}(t)$ represents the membrane potential, $I(t)$ is the external input current of the cell, ${s}$ represents the spike train. Moreover, $r$ is the membrane resistance and $\tau$ is the time constant. The firing of a spike is triggered when $u(t)$ reaches the threshold $U_{thr}$. 

\begin{figure}[h]
    \centering
    \includegraphics[width=1\linewidth]{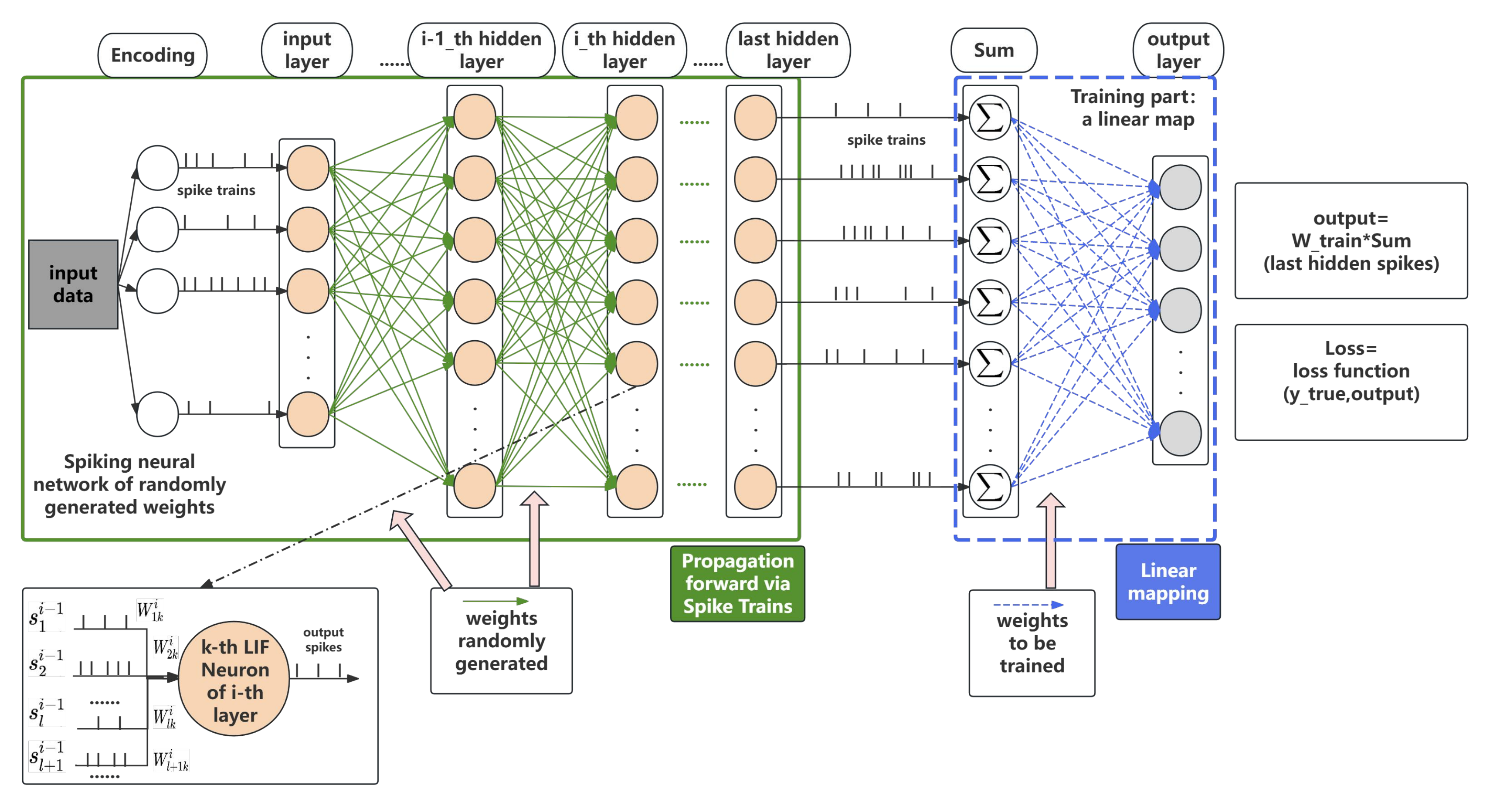}
    \caption{The composition of the SNN and the process of information propagation are as follows. The yellow neurons represent the Leaky Integrate-and-Fire (LIF) neuron model. The green connection weights are randomly generated and remain fixed (represented as green solid lines). In the last layer, the spike signals are accumulated and then undergo a linear mapping as the output (represented as blue dashed lines). The network task is completed by training this linear mapping}
    \label{fig:network}
\end{figure}

To form a multi-layer artificial neural network with spiking neurons, a general architecture is considered in this work, as sketched in Fig.\ref{fig:network}.  In this network, there are several hidden layers between the input layer and the output layer. The dynamics of the neurons in the $i$-th hidden layer are updated by the discrete version of LIF model as follows: 
\begin{equation}\label{discreteLIF}
    \left\{
    \begin{array}{l}
    \bm{u}^i(t+1) = \beta \bm{u}^i(t)+W^{i}\mathbf{s}^{i-1}(t)-{u}_{thr}\mathbf{s}^i(t) \\
    {s}^i_j(t) = 1,{\rm if}\:{u}^i_j(t)>\mathbf{u}_{thr}\\
    {s}^i_j(t) = 0,{\rm otherwise}\\
    \end{array}
    \right.
\end{equation}
where $\bm{u}^i=[u^i_1,u^i_2,\dots,u^i_n]^T$ is the state vector for the neurons of the $i$-th layer, $\bm{s}^i=[s^i_1,s^i_2,\dots,s^i_n]^T$ is the spike train vector of the $i$-th layer, and $W^i=\{W^i_{jk}\}$ is the weight matrix from $(i-1)$-th layer to the $i$-th layer, and $1-\beta$ represents the leaky rate. 

In traditional multilayer neural network frameworks, all weights in the network need to be adjusted through some form of training, in order to realize the approximating ability of the network. A key idea is to compute a loss function at the output layer and then use the chain rule to back-propagate the loss gradients with respect to the weights in each layer, thereby adjusting the weights in the direction of the negative gradient to complete the training process.  However, due to the discontinuous nature of spike firing in spiking neural networks (SNNs), gradient-based training methods cannot be directly applied. Even when surrogate gradient methods are introduced, issues such as vanishing or exploding gradients may still arise.

To address this, we propose a framework of randomized SNNs, known as RanSNN. In this framework, backpropagation of the loss function is restricted to only after the final layer of spiking neurons, while all preceding weights are randomly generated from a certain distribution and kept fixed (as shown by the solid lines in Fig.\ref{fig:network}). With such a setup, the trainable weights (indicated by dashed lines in Fig.\ref{fig:network}) can be optimized using the classical backpropagation algorithm. In particular, if the trainable weights are restricted to the final single layer, simple and effective linear minimization algorithms can be adopted.

The rationale behind this idea lies in the principle that the representation ability of the network is obtained from projecting input signal into a high-dimensional hidden space, where a series of nonlinear transformations are performed for tasks such as feature representation and information processing. The final output layer is responsible for extracting the processed information from the high-dimensional space back to a low-dimensional space. Numerous studies have shown that as long as the dimensionality of the hidden space is sufficiently high and the network depth is adequate, randomly assigned hidden layer connections can still provide enough representational and computational capacity \cite{jaeger2004harnessing,huang2004extreme}. Therefore, training only the output layer connections can still achieve the intended functionality of the neural network, significantly reducing the complexity of training while largely preserving the approximation capability of the network.



Next, we will introduce in detail how the network processes the input data and how it is trained.
Before the input layer, the input data is first encoded into spiking trains with a Poisson encoding schema.  That is, the input is expanded and, after normalizing the input data, Poisson encoding is carried out according to the normalized data to generate spike trains over a predefined time length of time steps. 
Then, as shown in Fig.\ref{fig:network}, the weights between the hidden layers composed of Leaky Integrate-and-Fire (LIF) neurons are randomly generated according to a specific distribution and then fixed. Thus, the hidden layers are responsible for the forward propagation of signals in the dynamic behavior described above. Until the last hidden layer, we perform some special processing on the spike trains generated by the last hidden layer. We sum up the spike train generated by each neuron to obtain the total number of pulse signals released by each neuron within our decoding time steps.  Then a loss function is designed according to the specific task of the neural network and calculated based on the output result and the target value $y_{true}$, and this loss is used to train the weights $W_{train}$ that are trainable.  

It is worth noting that this is a general architecture where several elements are flexible.  In Fig.\ref{fig:network}, the trainable weights are illustrated between the last hidden layer and the output layer. However, weight training is not restricted to the final layer-it can be extended to multiple layers, provided that no spiking neuron layers are involved. Moreover the training algorithm could vary according to the specific setting. In this work, only the weights between the last hidden layer and the output layer are trained and thus the simple linear minimization methods is enough to solve it,  while for situations when multiple layers needs training,  the classic back-propagation methods could be adopted.

\subsection{Benchmark Validations}
To validate the proposed RanSNN framework, we take the MNIST task  as a benchmark. In this task, each handwritten digit image is a  28×28-pixel grayscale image, and the corresponding label is the category of the digit, ranging from 0 to 9. We first expand the MNIST image into 784 input neurons, where each neuron represents a pixel in the image. Then generate spike signals over a time step of 25 using the normalized pixel values as the Poisson intensity. In this way, an image can be encoded into 784 input spike trains. To make the illustration as well as the discussions simple and clear, we setup the hidden layers with depth $1$ and width $2000$, i.e., there is one hidden layer of $2000$ LIF neurons in the architecture of Fig.\ref{fig:network}.

After completing the encoding of the input signal and the setup of the network structure, we randomly generate the connection weights between the input layer and the hidden layer (\textbf{W} in Equation \ref{discreteLIF}) by the uniform distribution ${\rm U}(-a,a)$ where $a$ is set as $\sqrt{6/784}$.

The loss function is chosen as the cross-entroy which is generally used in the classification task. Specifically, the cross-entropy we calculate is:
\begin{equation}
    H(y_{true}, \textbf{output}) = -\sum (y_{true} * \log(\textbf{output}))
\end{equation}

For the MNIST dataset, we select 400 batches as the training set and 50 batches as the test set with 128 images in each batch. We use pytorch's built-in 
Adam optimizer for training in each iteration\cite{10242251}.

To examine the universality and stability of our training algorithm across tasks of varying difficulty levels, we have selected four tasks with different difficulty levels and focuses to verify our algorithm respectively.

Our task setup is as follows. Keeping the input encoding method and the network structures of the input layer and the hidden layer constant, we respectively perform recognition on four different datasets. They are the classic handwritten digit dataset MNIST, the Fashion - MNIST (fMNIST) dataset which consists of fashion items and has greater intra - class differences in data\cite{xiao2017fashion}, the Kuzushiji - MNIST(kMNIST) dataset composed of handwritten Japanese kana with unique writing styles, and the Extended MNIST(eMNIST). The Extended MNIST has a large data scale and rich categories, including 62 character categories covering numbers 0 - 9, uppercase letters A - Z, and lowercase letters a - z\cite{cohen2017emnist}. Among them, the output dimensions of the first three tasks are all 10, and the output dimension of the last task is 62.

The setup of the training set and the test set remains consistent with the above. There are 400 batches of 128 images used as the training set and 50 batches of 128 images used as the test set. 

Considering that our algorithm offers a high degree of freedom in network architecture, we take the surrogate gradient (SG) method as the comparison base to illustrate the advantages and disadvantages of our network.  While keeping the network structure unchanged, instead of summing up the connections between the hidden layer and the output layer, they are made consistent with those between the input layer and the hidden layer with SG method. For each step of the spike in the output layer, we expect that the neuron corresponding to the correctly recognized result emits the largest number of spikes. Therefore, we calculate the cross-entropy between the membrane potential at each step and the target, and then sum up the cross-entropies at each moment to use as the final loss function for training. The details for the SG method is provided in Sec.\ref{detail}. 

\begin{figure}[h]
\centering  
\subfigure[]{
\label{Fig.sub.tr1}
\includegraphics[width=0.4\textwidth]{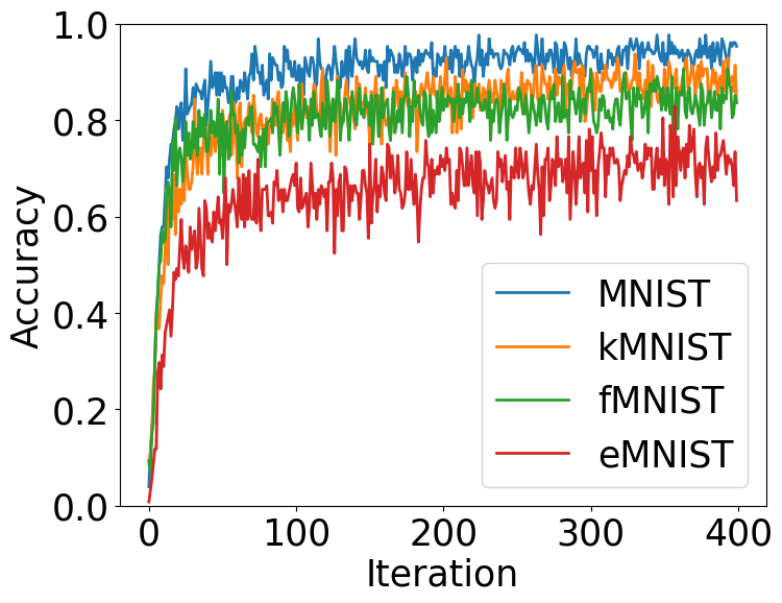}}
\subfigure[]{
\label{Fig.sub.tr2}
\includegraphics[width=0.4\textwidth]{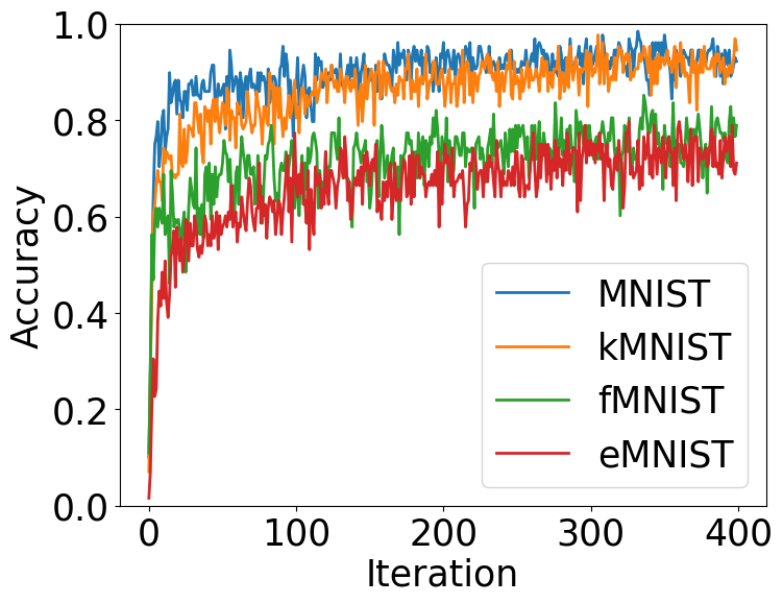}}

\subfigure[]{
\label{Fig.sub.ts1}
\includegraphics[width=0.4\textwidth]{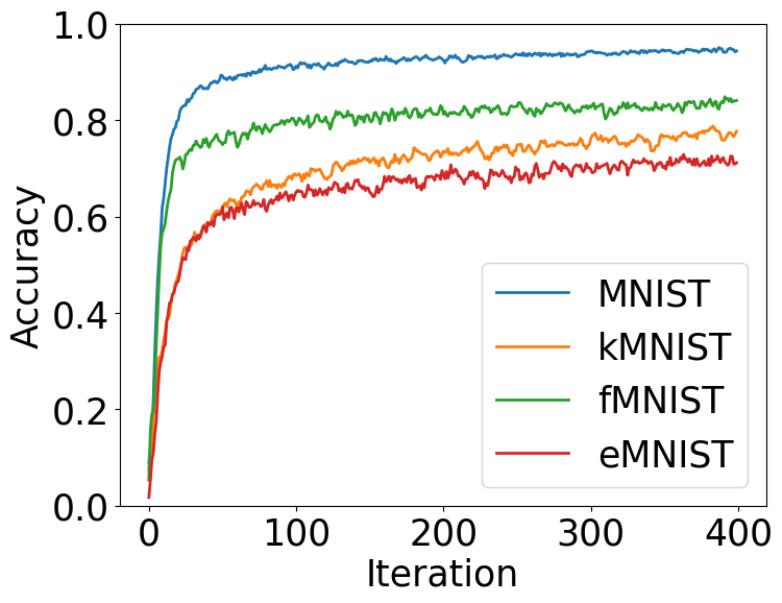}}
\subfigure[]{
\label{Fig.sub.ts2}
\includegraphics[width=0.4\textwidth]{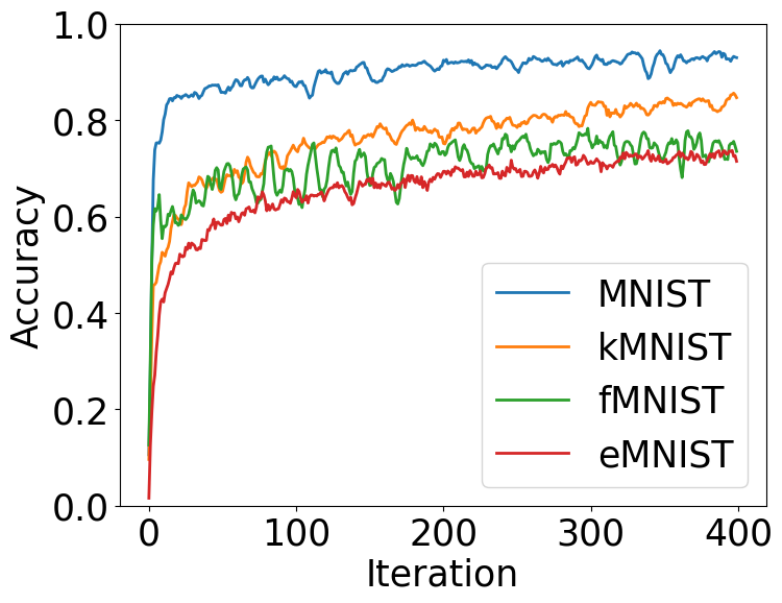}}

\subfigure[]{
\label{Fig.sub.l1}
\includegraphics[width=0.4\textwidth]{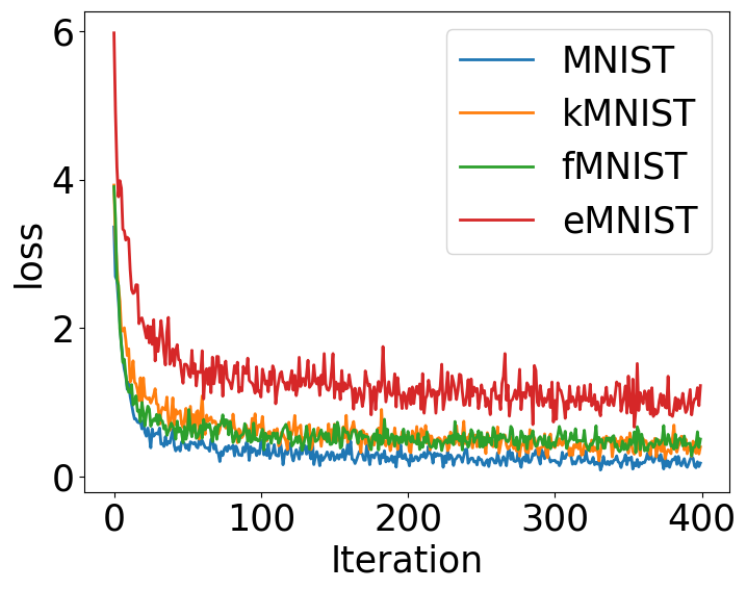}}
\subfigure[]{
\label{Fig.sub.l2}
\includegraphics[width=0.4\textwidth]{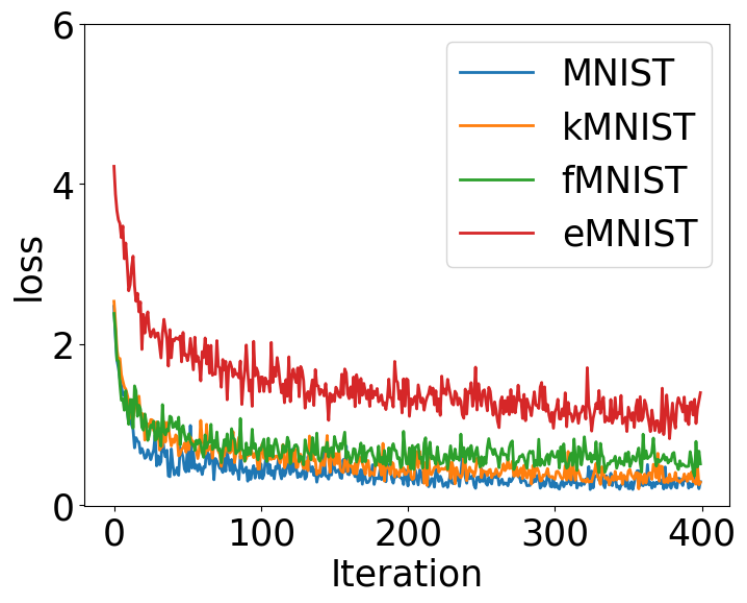}}
\caption{Performance of SG and RanSNN: the convergence of train accuracy, test accuracy, and loss as iteration increases. (a)(c)(e) for RanSNN and (b)(d)(f) for SG respectively.  }
\label{r}
\end{figure}

The performance of both methods are depicted in Fig.\ref{r},  where the train accuracy, test accuracy as well as the loss are shown. It is shown that compared with the SG method, the RanSNN  can also stably achieve similar performance. While the performance is similar, it is noted that due to the relatively simple training tasks, the RanSNN has a significant advantage in training efficiency compared to the traditional surrogate gradient method. In Table \ref{table1}, we compare the test accuracies and training times of the four tasks. It can be seen that since RanSNN only needs to complete a linear optimization, it can achieve performance close to that of the surrogate gradient method but with an extremely shorter time.

\begin{table}[h]
  \begin{center}
    \caption{Accuracy and training time of the two methods}\label{table1}
    \begin{tabular}{l|c|c|c|r}
        \toprule
        Dataset & MNIST & fMNIST & kMNIST& eMNIST\\
        \midrule
        training time(SG) & 73.77s& 70.54s& 72.04s& 76.80s\\
        training time(RanSNN) & 0.6699s& 0.8417s& 0.8601s& 1.1732s\\
        accuracy (SG) & 93.41\%& 72.52\%& 81.84\%& 72.42\%\\
        accuracy (RanSNN) & 92.62\%& 85.12\%& 70.77\%& 70.16\%\\
        \bottomrule
    \end{tabular}
  \end{center}
\end{table}

\section{Discussions}\label{sec:discussions}
\subsection{Analysis of hyper-parameters}
In this section, we will discuss the influence of the hyper-parameters such as the leaky rate $\beta$, the number of neurons in the hidden layer, and the number of time steps. These hyper-parameters play a decisive role in determining the outcome of the entire experiment. 

In the following experiments, by default, the value of $\beta$ is set to 0.95, the number of hidden neurons is 2000, and the number of time steps is 25. When conducting experiments on different parameters, we keep the other two parameters at their default values, and also complete the four tasks mentioned in the previous text.

\begin{figure}[h]
\centering
\subfigure[]{
\label{Fig.sub.h1}
\includegraphics[width=0.32\textwidth]{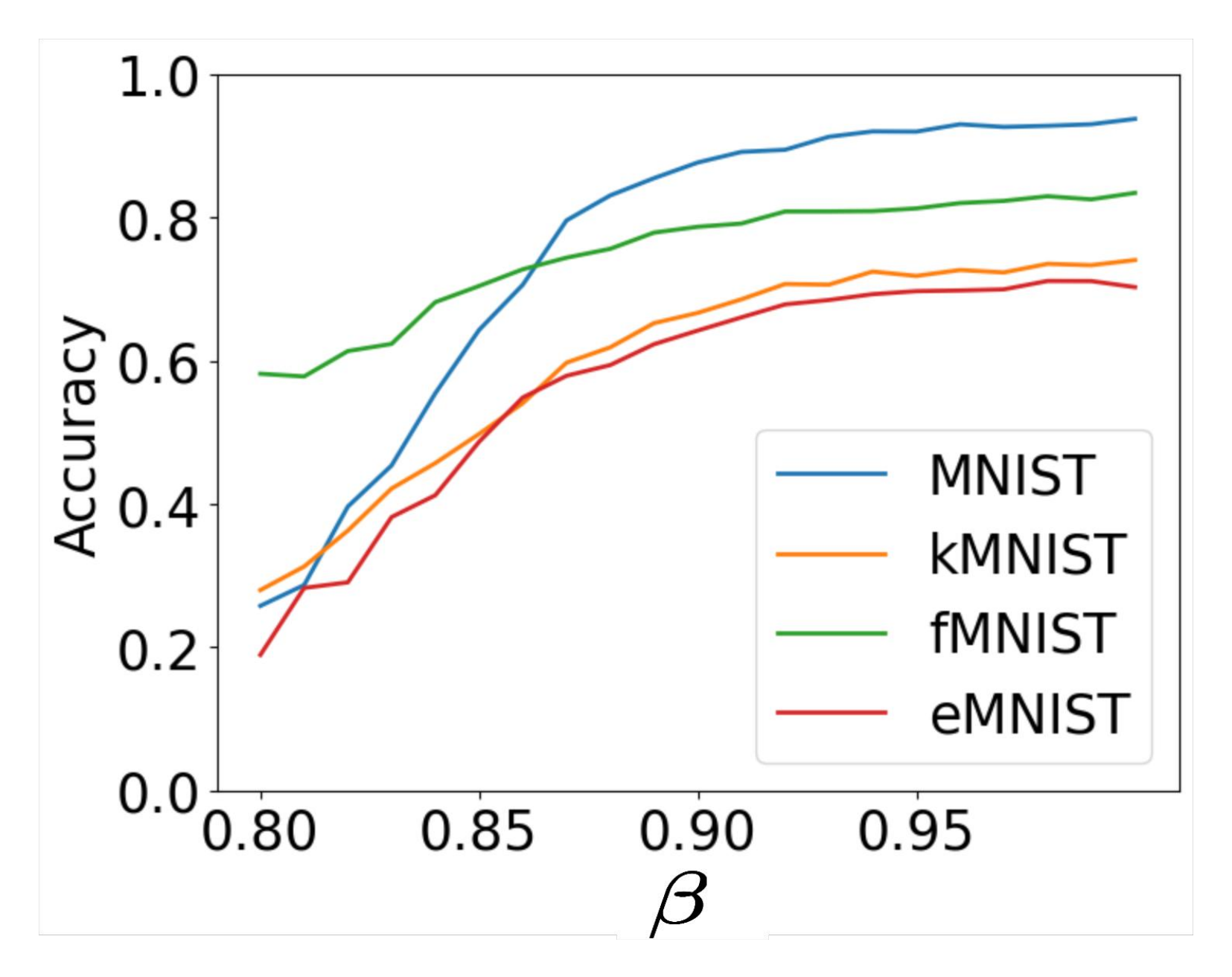}}
\subfigure[]{
\label{Fig.sub.h2}
\includegraphics[width=0.32\textwidth]{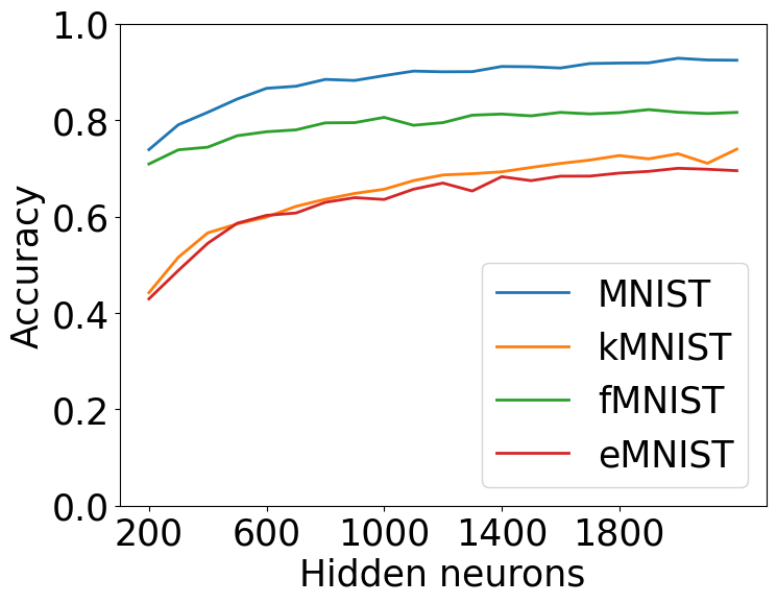}}
\subfigure[]{
\label{Fig.sub.h3}
\includegraphics[width=0.32\textwidth]{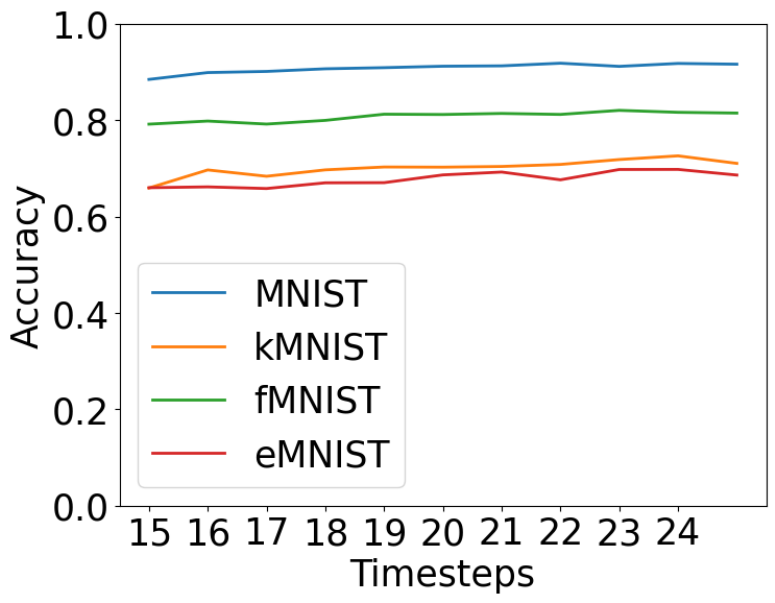}}
\caption{The impact of hyper-parameters on the results. (a)The leakage parameter $\beta$;  (b)The number of neurons in the hidden layer; (c)The length of the spike train time. }
\label{h}
\end{figure}

First of all, $\beta$, as a hyperparameter of the discretized Leaky Integrate-and-Fire (LIF) neuron, holds a crucial position.  The leaky characteristic represented by $1-\beta$ in the LIF neuron essentially describes the decay of the membrane potential over time when no external input is received. When the degree of decay is too large, it is difficult for the membrane potential of the neuron to accumulate to the threshold for triggering the release of the spike signal, thus causing the neuron to fail to release the spike signal. Thus, the leakage rate of neurons needs to be controlled at a relatively low level to ensure that spiking neurons can discharge and a large $\beta$ is preferred respectively, as shown in Fig.\ref{h}(a).

Then the width of the hidden layer, i.e., the number of hidden-neurons is another important hyper-parameter. To study the choice of neuron number, we adopt a learning method similar to the Extreme Learning Machine (ELM)\cite{HUANG2006489}. In this learning mode, there is a close and complex relationship between the memory ability of the network and the number of neurons in the hidden layer. By continuously adjusting the number of neurons in the hidden layer and observing the performance of the network on different datasets, we try to find the number of neurons in the hidden layer that can achieve the best balance between energy consumption and performance. Clearly we found that there is a significant positive correlation between the accuracy rate and the number of neurons in the hidden layer, as shown in Fig.\ref{h}(b).

Finally, as mentioned earlier, when we input network data, such as an image, into the network for processing, we need to perform an encode operation on it first. The purpose of encoding is to convert the original data into the form of spike sequences suitable for processing by the SNN. At this step, a crucial question emerges: what scale of spike sequence should we encode the original input data into so as to retain the features of the original input signal to the greatest extent? To study this problem, we use different values of time steps and conduct a comparative analysis of the degree to which the network retains the features of the original input signal and the final processing effect under different circumstances. As shown in Fig.\ref{h}(c), it can be found that the increase in time steps cannot steadily improve the performance of the network.

\subsection{Analysis of random weights}
In this section, we specifically focus on one question: how much do randomly generated weights affect the results? Considering that random generation is a broad concept, we will next concentrate on two common random generation methods: generation according to a uniform distribution and generation according to a normal distribution.

\begin{figure}[h]
\centering  
\subfigure[]{
\label{Fig.sub.u1}
\includegraphics[width=0.4\textwidth]{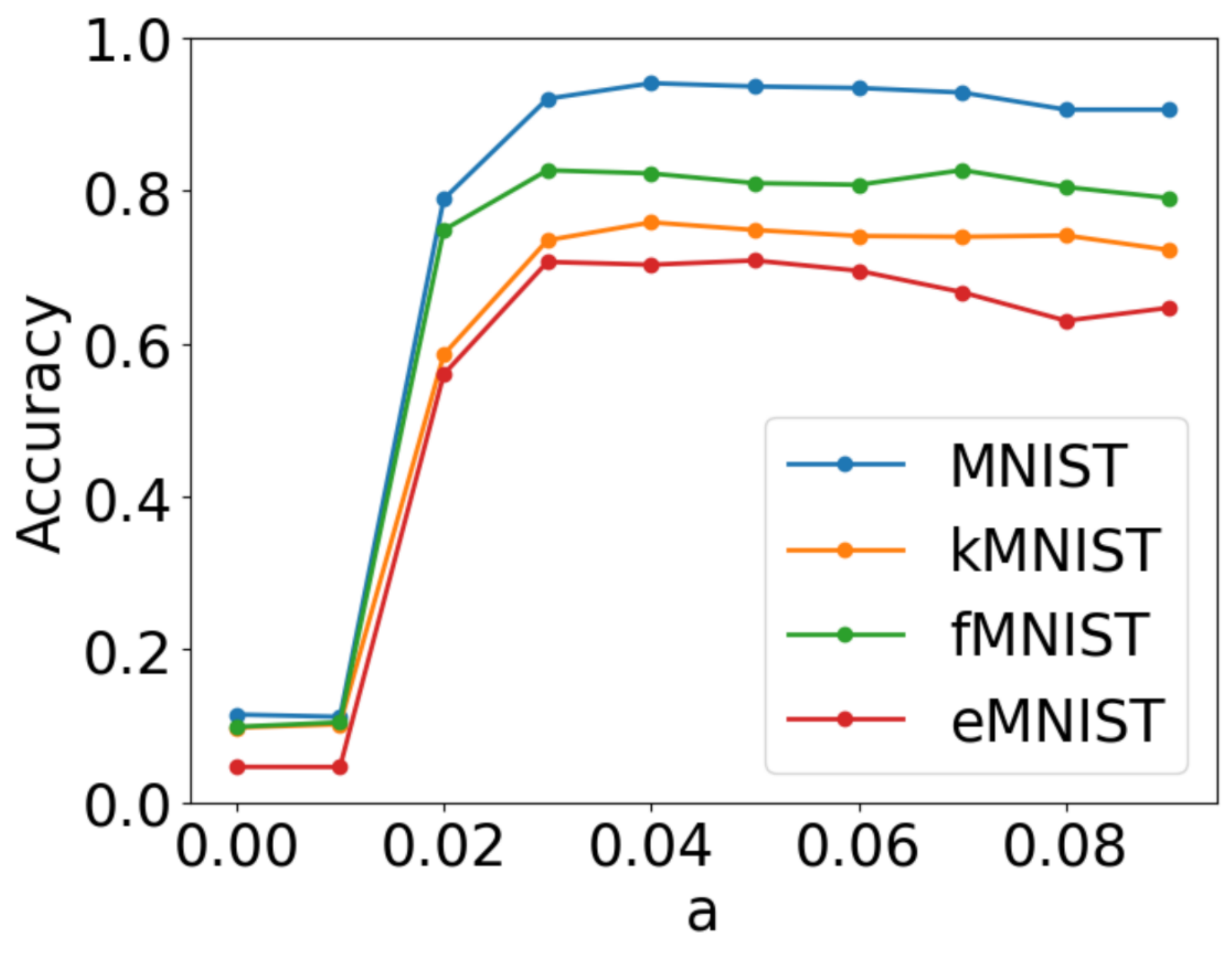}}
\subfigure[]{
\label{Fig.sub.u2}
\includegraphics[width=0.4\textwidth]{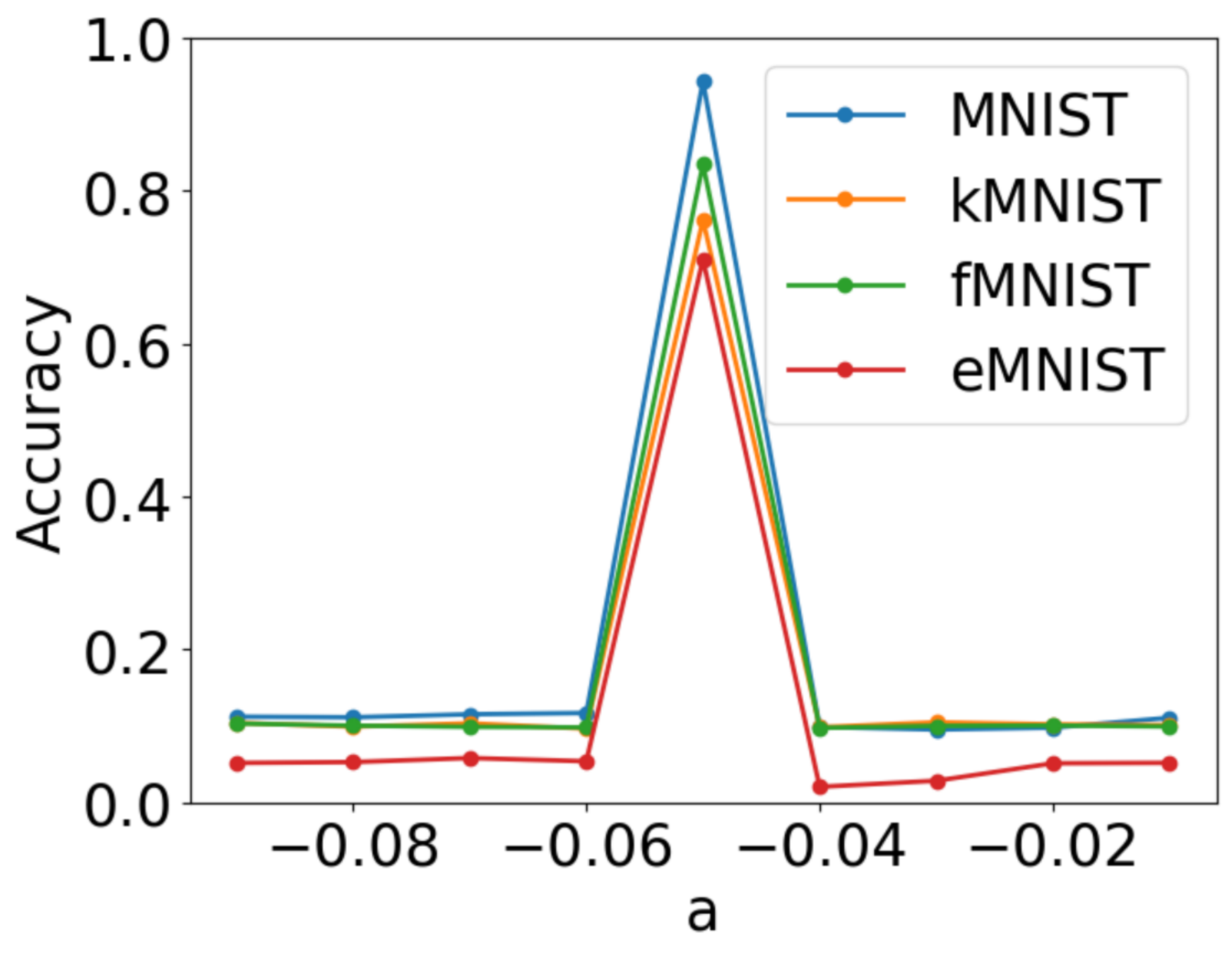}}
\caption{Performance under different random weights generation by uniform distributions:(a) ${\rm U}(-a,a)$; (b)${\rm U}(a,a+0.1)$} 
\label{u}
\end{figure}

As for the uniform distribution, we carry out tests under two types of uniform distributions: one is centered around $0$ while the other is biased, as shown in Fig.\ref{u}.  It can be seen that for RanSNN to maintain its effectiveness, the uniform distribution needs to be centered around 0 and have a range larger than a threshold value (in this work the threshold is around $0.02$). From a biological logic perspective, it means that biological neurons need both inhibitive and active inputs to process the information.  

\begin{figure}[h]
\centering  
\subfigure[]{
\label{Fig.sub.n1}
\includegraphics[width=0.4\textwidth]{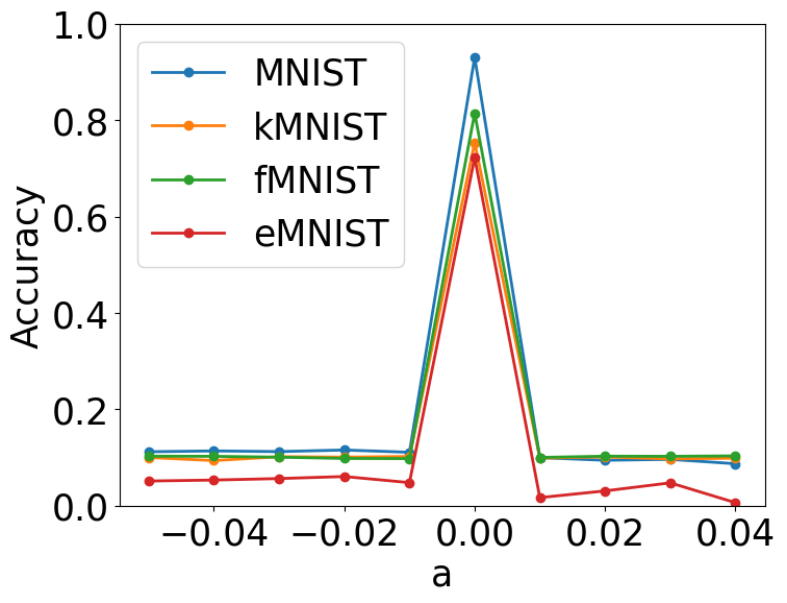}}
\subfigure[]{
\label{Fig.sub.n2}
\includegraphics[width=0.4\textwidth]{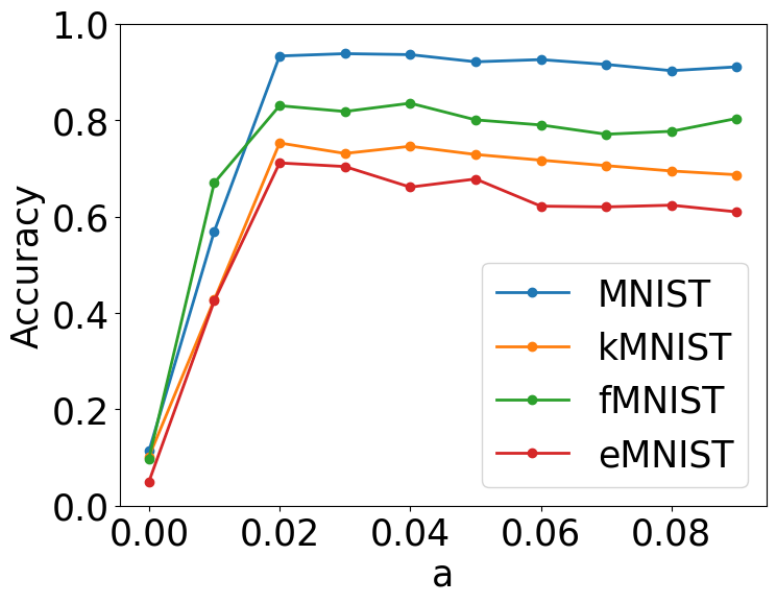}}
\caption{Performance under different random weights generation by normal distributions : (a) normal distribution ${\rm N}(a,1)$; (b) normal distribution ${\rm N}(0,a)$.}
\label{n}
\end{figure}

Then we consider the random weights generated from normal distributions. We also consider two types: ${\rm N}(a,1)$ and ${\rm N}(0,a)$. It is clear to see from Fig.\ref{n}(a) that the requirement of keeping the mean at 0. The results in Fig. \ref{n}(b) imply that the standard deviation also needs to be greater than a threshold to ensure that the network matches the scale of its input. These results confirm that both uniform and normal distributions can work in the RanSNN framework, but $0$ expectation is essential and the strength should be larger than a threshold. 

\section{Limitations of our method}
In this section, we mainly discuss the limitations and deficiencies of the RanSNN framework proposed in this work.

Firstly, this framework cannot be generalized into some networks where the connection weights have specific meanings. For example, in a convolutional neural network, the connection weights are convolution kernels. The data of the convolution kernels determines the processing method for image features and thus should be trained rather than randomized. 

The second point is regarding the processing of the last layer of the network. We summed the outputs of the last hidden layer and then trained a simple linear mapping. This approach may lead to the loss of original features when dealing with more complex tasks, making it difficult to achieve satisfactory training results.

\section{Summary}
In this research, a novel architecture for spiking neural networks, RanSNN, is proposed. To thoroughly explore the performance of this architecture, a series of numerical experiments are conducted. The experimental results validate that this architecture exhibits effectiveness in lightweight network scenarios, fully highlighting its significant advantages of high training efficiency and a concise training process.

It should not be overlooked that there are still certain limitations and deficiencies in the current training architecture. However, this architecture has opened up entirely new research ideas and directions in the field of spiking neural network training. Particularly, this training algorithm demonstrates a high degree of flexibility in terms of network structure adaptation and task setting, and exhibits certain characteristics of universality. This implies that it has potential application values and exploration space under different application scenarios and research requirements, and is expected to provide strong support for the further development of spiking neural networks.

A prominent characteristic of spiking neural networks lies in the fact that their distinctive information transmission mode endows the network with high energy efficiency. Although the RanSNN framework proposed in this study may have some room for improvement in terms of task execution effectiveness, it has achieved a significant breakthrough in training efficiency. This efficiency advantage not only aligns with the inherent energy efficiency characteristics of spiking neural networks but also largely demonstrates the potential and value of this method in practical applications. Through an efficient training process, this method can achieve similar task goals in a shorter time, thus providing a more feasible solution for large scale data processing and applications with high real time requirements.

\section{Implementation details}
\begin{table}[h]
  \begin{center}
    \caption{Environment configuration}
    \begin{tabular}{l|r}
        \toprule
        item & configuration\\
        \midrule
        Operation system & windows11\\
        CPU & Intel Core i9-13980HX Processor\\
        GPU & Intel UHD Graphics for 13th Gen Intel Processors\\
        python & 3.8\\
        \bottomrule
    \end{tabular}
  \end{center}
\end{table}

\newpage
\bibliographystyle{unsrt}
\bibliography{references}


\appendix

\section{Technical Appendices and Supplementary Material}\label{detail}
\subsection{The Surragate Gradient method}
\begin{figure}[h]
    \centering
    \includegraphics[width=0.5\linewidth]{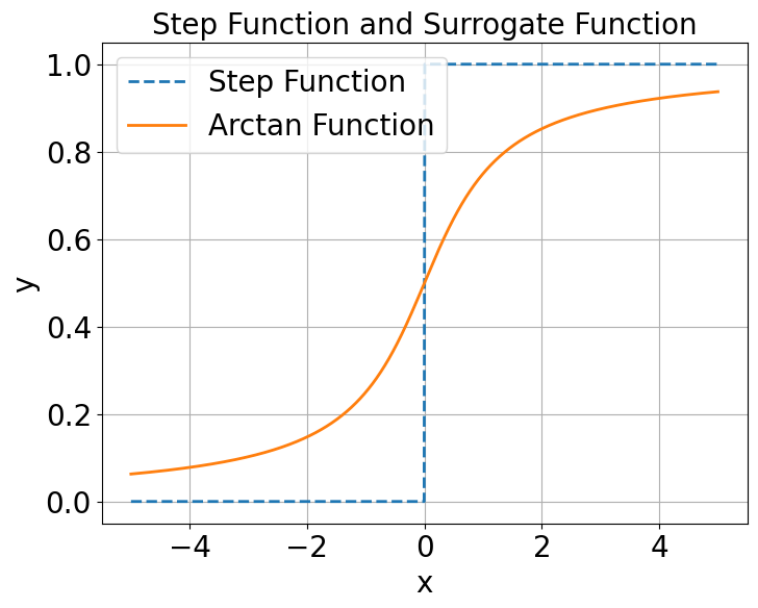}
    \caption{Surrogate function}
    \label{fig:sf}
\end{figure}

In Section 2, we compare and discuss the proposed RanSNN with the SG method. Here, when we use the surrogate gradient method, we keep the network structure unchanged and remove the summation layer. We directly connect to the output layer, and all the weights need to be trained. We make the following stipulations: when we calculate the derivative, we replace the jump function with the inverse tangent function shown in Fig. \ref{fig:sf} to implement gradient descent. At each time step, we expect the "correct" neurons to respond. Therefore, within each time step of the output layer, we calculate the cross-entropy between the current membrane potential and the target label, and then accumulate these cross-entropies over time to serve as the loss function. To compare with the RanSNN, the loss shown in Fig.\ref{r}(f) is averaged by the time steps. 

\section{Dataset download}
The download address of the dataset required for the article is given in the following table. 
\begin{table}[h]
  \begin{center}
    \caption{Dataset download address}\label{table2}
    \begin{tabular}{l|r}
        \toprule
        Dataset & \\
        \midrule
        MNIST & https://yann.lecun.com/exdb/mnist/\\
        fMNIST & https://github.com/zalandoresearch/fashion-mnist\\
        kMNIST & https://github.com/rois-codh/kmnist\\
        eMNIST & https://docs.pytorch.org/vision/main/generated/torchvision.datasets.EMNIST.html\\
        \bottomrule
    \end{tabular}
  \end{center}
\end{table}


\newpage
\section*{NeurIPS Paper Checklist}

\begin{enumerate}

\item {\bf Claims}
    \item[] Question: Do the main claims made in the abstract and introduction accurately reflect the paper's contributions and scope?
    \item[] Answer: \answerYes{}
    \item[] Justification: In this work we propose a new framework of spiking neural networks. Comparing with traditional training methods for spiking neural networks, it greatly improves the training efficiency and also ensures good network performance.
    \item[] Guidelines:
    \begin{itemize}
        \item The answer NA means that the abstract and introduction do not include the claims made in the paper.
        \item The abstract and/or introduction should clearly state the claims made, including the contributions made in the paper and important assumptions and limitations. A No or NA answer to this question will not be perceived well by the reviewers.
        \item The claims made should match theoretical and experimental results, and reflect how much the results can be expected to generalize to other settings.
        \item It is fine to include aspirational goals as motivation as long as it is clear that these goals are not attained by the paper.
    \end{itemize}

\item {\bf Limitations}
    \item[] Question: Does the paper discuss the limitations of the work performed by the authors?
    \item[] Answer: \answerYes{}
    \item[] Justification: In section 3, we discuss about two main limitations of our method.
    \item[] Guidelines:
    \begin{itemize}
        \item The answer NA means that the paper has no limitation while the answer No means that the paper has limitations, but those are not discussed in the paper.
        \item The authors are encouraged to create a separate "Limitations" section in their paper.
        \item The paper should point out any strong assumptions and how robust the results are to violations of these assumptions (e.g., independence assumptions, noiseless settings, model well-specification, asymptotic approximations only holding locally). The authors should reflect on how these assumptions might be violated in practice and what the implications would be.
        \item The authors should reflect on the scope of the claims made, e.g., if the approach was only tested on a few datasets or with a few runs. In general, empirical results often depend on implicit assumptions, which should be articulated.
        \item The authors should reflect on the factors that influence the performance of the approach. For example, a facial recognition algorithm may perform poorly when image resolution is low or images are taken in low lighting. Or a speech-to-text system might not be used reliably to provide closed captions for online lectures because it fails to handle technical jargon.
        \item The authors should discuss the computational efficiency of the proposed algorithms and how they scale with dataset size.
        \item If applicable, the authors should discuss possible limitations of their approach to address problems of privacy and fairness.
        \item While the authors might fear that complete honesty about limitations might be used by reviewers as grounds for rejection, a worse outcome might be that reviewers discover limitations that aren't acknowledged in the paper. The authors should use their best judgment and recognize that individual actions in favor of transparency play an important role in developing norms that preserve the integrity of the community. Reviewers will be specifically instructed to not penalize honesty concerning limitations.
    \end{itemize}

\item {\bf Theory assumptions and proofs}
    \item[] Question: For each theoretical result, does the paper provide the full set of assumptions and a complete (and correct) proof?
    \item[] Answer: \answerNA{}
    \item[] Justification: The main work of this paper is to propose a new SNN framework design, which does not involve theoretical results.
    \item[] Guidelines:
    \begin{itemize}
        \item The answer NA means that the paper does not include theoretical results.
        \item All the theorems, formulas, and proofs in the paper should be numbered and cross-referenced.
        \item All assumptions should be clearly stated or referenced in the statement of any theorems.
        \item The proofs can either appear in the main paper or the supplemental material, but if they appear in the supplemental material, the authors are encouraged to provide a short proof sketch to provide intuition.
        \item Inversely, any informal proof provided in the core of the paper should be complemented by formal proofs provided in appendix or supplemental material.
        \item Theorems and Lemmas that the proof relies upon should be properly referenced.
    \end{itemize}

    \item {\bf Experimental result reproducibility}
    \item[] Question: Does the paper fully disclose all the information needed to reproduce the main experimental results of the paper to the extent that it affects the main claims and/or conclusions of the paper (regardless of whether the code and data are provided or not)?
    \item[] Answer: \answerYes{}
    \item[] Justification: The schematic diagram of the network structure and its textual description in Section \ref{sec:method} of this paper fully explain the implementation process of the algorithm, and the reader can reproduce the experimental results through the content of this section.
    \item[] Guidelines:
    \begin{itemize}
        \item The answer NA means that the paper does not include experiments.
        \item If the paper includes experiments, a No answer to this question will not be perceived well by the reviewers: Making the paper reproducible is important, regardless of whether the code and data are provided or not.
        \item If the contribution is a dataset and/or model, the authors should describe the steps taken to make their results reproducible or verifiable.
        \item Depending on the contribution, reproducibility can be accomplished in various ways. For example, if the contribution is a novel architecture, describing the architecture fully might suffice, or if the contribution is a specific model and empirical evaluation, it may be necessary to either make it possible for others to replicate the model with the same dataset, or provide access to the model. In general. releasing code and data is often one good way to accomplish this, but reproducibility can also be provided via detailed instructions for how to replicate the results, access to a hosted model (e.g., in the case of a large language model), releasing of a model checkpoint, or other means that are appropriate to the research performed.
        \item While NeurIPS does not require releasing code, the conference does require all submissions to provide some reasonable avenue for reproducibility, which may depend on the nature of the contribution. For example
        \begin{enumerate}
            \item If the contribution is primarily a new algorithm, the paper should make it clear how to reproduce that algorithm.
            \item If the contribution is primarily a new model architecture, the paper should describe the architecture clearly and fully.
            \item If the contribution is a new model (e.g., a large language model), then there should either be a way to access this model for reproducing the results or a way to reproduce the model (e.g., with an open-source dataset or instructions for how to construct the dataset).
            \item We recognize that reproducibility may be tricky in some cases, in which case authors are welcome to describe the particular way they provide for reproducibility. In the case of closed-source models, it may be that access to the model is limited in some way (e.g., to registered users), but it should be possible for other researchers to have some path to reproducing or verifying the results.
        \end{enumerate}
    \end{itemize}

\item {\bf Open access to data and code}
    \item[] Question: Does the paper provide open access to the data and code, with sufficient instructions to faithfully reproduce the main experimental results, as described in supplemental material?
    \item[] Answer: \answerYes{}
    \item[] Justification: All the codes and instructions can be referred to  \url{https://github.com/Asd0731/Spiking-Neural-Networks-with-Random-Network-Architecture}
    \item[] Guidelines:
    \begin{itemize}
        \item The answer NA means that paper does not include experiments requiring code.
        \item Please see the NeurIPS code and data submission guidelines (\url{https://nips.cc/public/guides/CodeSubmissionPolicy}) for more details.
        \item While we encourage the release of code and data, we understand that this might not be possible, so “No” is an acceptable answer. Papers cannot be rejected simply for not including code, unless this is central to the contribution (e.g., for a new open-source benchmark).
        \item The instructions should contain the exact command and environment needed to run to reproduce the results. See the NeurIPS code and data submission guidelines (\url{https://nips.cc/public/guides/CodeSubmissionPolicy}) for more details.
        \item The authors should provide instructions on data access and preparation, including how to access the raw data, preprocessed data, intermediate data, and generated data, etc.
        \item The authors should provide scripts to reproduce all experimental results for the new proposed method and baselines. If only a subset of experiments are reproducible, they should state which ones are omitted from the script and why.
        \item At submission time, to preserve anonymity, the authors should release anonymized versions (if applicable).
        \item Providing as much information as possible in supplemental material (appended to the paper) is recommended, but including URLs to data and code is permitted.
    \end{itemize}

\item {\bf Experimental setting/details}
    \item[] Question: Does the paper specify all the training and test details (e.g., data splits, hyperparameters, how they were chosen, type of optimizer, etc.) necessary to understand the results?
    \item[] Answer:\answerYes{}
    \item[] Justification: Later in Section 2.1, we describe in detail how the training data enters the network, what parts of the network need to be trained, using cross-entropy as the loss function and clarifying the choice of optimizer.
    \item[] Guidelines:
    \begin{itemize}
        \item The answer NA means that the paper does not include experiments.
        \item The experimental setting should be presented in the core of the paper to a level of detail that is necessary to appreciate the results and make sense of them.
        \item The full details can be provided either with the code, in appendix, or as supplemental material.
    \end{itemize}

\item {\bf Experiment statistical significance}
    \item[] Question: Does the paper report error bars suitably and correctly defined or other appropriate information about the statistical significance of the experiments?
    \item[] Answer: \answerYes{}
    \item[] Justification: In section \ref{sec:discussions}, we discuss the effect of parameter fluctuations and randomly generated distributions on the results
    \item[] Guidelines:
    \begin{itemize}
        \item The answer NA means that the paper does not include experiments.
        \item The authors should answer "Yes" if the results are accompanied by error bars, confidence intervals, or statistical significance tests, at least for the experiments that support the main claims of the paper.
        \item The factors of variability that the error bars are capturing should be clearly stated (for example, train/test split, initialization, random drawing of some parameter, or overall run with given experimental conditions).
        \item The method for calculating the error bars should be explained (closed form formula, call to a library function, bootstrap, etc.)
        \item The assumptions made should be given (e.g., Normally distributed errors).
        \item It should be clear whether the error bar is the standard deviation or the standard error of the mean.
        \item It is OK to report 1-sigma error bars, but one should state it. The authors should preferably report a 2-sigma error bar than state that they have a 96\% CI, if the hypothesis of Normality of errors is not verified.
        \item For asymmetric distributions, the authors should be careful not to show in tables or figures symmetric error bars that would yield results that are out of range (e.g. negative error rates).
        \item If error bars are reported in tables or plots, The authors should explain in the text how they were calculated and reference the corresponding figures or tables in the text.
    \end{itemize}

\item {\bf Experiments compute resources}
    \item[] Question: For each experiment, does the paper provide sufficient information on the computer resources (type of compute workers, memory, time of execution) needed to reproduce the experiments?
    \item[] Answer: \answerYes{}
    \item[] Justification: In Section 5, we list the experimental environment and experimental equipment.
    \item[] Guidelines:
    \begin{itemize}
        \item The answer NA means that the paper does not include experiments.
        \item The paper should indicate the type of compute workers CPU or GPU, internal cluster, or cloud provider, including relevant memory and storage.
        \item The paper should provide the amount of compute required for each of the individual experimental runs as well as estimate the total compute.
        \item The paper should disclose whether the full research project required more compute than the experiments reported in the paper (e.g., preliminary or failed experiments that didn't make it into the paper).
    \end{itemize}

\item {\bf Code of ethics}
    \item[] Question: Does the research conducted in the paper conform, in every respect, with the NeurIPS Code of Ethics \url{https://neurips.cc/public/EthicsGuidelines}?
    \item[] Answer: \answerYes{}
    \item[] Justification: This work is in accordance with NIPS ethical standards.
    \item[] Guidelines:
    \begin{itemize}
        \item The answer NA means that the authors have not reviewed the NeurIPS Code of Ethics.
        \item If the authors answer No, they should explain the special circumstances that require a deviation from the Code of Ethics.
        \item The authors should make sure to preserve anonymity (e.g., if there is a special consideration due to laws or regulations in their jurisdiction).
    \end{itemize}

\item {\bf Broader impacts}
    \item[] Question: Does the paper discuss both potential positive societal impacts and negative societal impacts of the work performed?
    \item[] Answer: \answerYes{}
    \item[] Justification: In the course of the discussion in Section 5, we describe the advantages and possible implications of the algorithm in this paper.
    \item[] Guidelines:
    \begin{itemize}
        \item The answer NA means that there is no societal impact of the work performed.
        \item If the authors answer NA or No, they should explain why their work has no societal impact or why the paper does not address societal impact.
        \item Examples of negative societal impacts include potential malicious or unintended uses (e.g., disinformation, generating fake profiles, surveillance), fairness considerations (e.g., deployment of technologies that could make decisions that unfairly impact specific groups), privacy considerations, and security considerations.
        \item The conference expects that many papers will be foundational research and not tied to particular applications, let alone deployments. However, if there is a direct path to any negative applications, the authors should point it out. For example, it is legitimate to point out that an improvement in the quality of generative models could be used to generate deepfakes for disinformation. On the other hand, it is not needed to point out that a generic algorithm for optimizing neural networks could enable people to train models that generate Deepfakes faster.
        \item The authors should consider possible harms that could arise when the technology is being used as intended and functioning correctly, harms that could arise when the technology is being used as intended but gives incorrect results, and harms following from (intentional or unintentional) misuse of the technology.
        \item If there are negative societal impacts, the authors could also discuss possible mitigation strategies (e.g., gated release of models, providing defenses in addition to attacks, mechanisms for monitoring misuse, mechanisms to monitor how a system learns from feedback over time, improving the efficiency and accessibility of ML).
    \end{itemize}

\item {\bf Safeguards}
    \item[] Question: Does the paper describe safeguards that have been put in place for responsible release of data or models that have a high risk for misuse (e.g., pretrained language models, image generators, or scraped datasets)?
    \item[] Answer: \answerNA{}.
    \item[] Justification: The paper poses no such risks.
    \item[] Guidelines:
    \begin{itemize}
        \item The answer NA means that the paper poses no such risks.
        \item Released models that have a high risk for misuse or dual-use should be released with necessary safeguards to allow for controlled use of the model, for example by requiring that users adhere to usage guidelines or restrictions to access the model or implementing safety filters.
        \item Datasets that have been scraped from the Internet could pose safety risks. The authors should describe how they avoided releasing unsafe images.
        \item We recognize that providing effective safeguards is challenging, and many papers do not require this, but we encourage authors to take this into account and make a best faith effort.
    \end{itemize}

\item {\bf Licenses for existing assets}
    \item[] Question: Are the creators or original owners of assets (e.g., code, data, models), used in the paper, properly credited and are the license and terms of use explicitly mentioned and properly respected?
    \item[] Answer: \answerYes{}
    \item[] Justification: The tools and datasets used in this paper are all marked with their sources.
    \item[] Guidelines:
    \begin{itemize}
        \item The answer NA means that the paper does not use existing assets.
        \item The authors should cite the original paper that produced the code package or dataset.
        \item The authors should state which version of the asset is used and, if possible, include a URL.
        \item The name of the license (e.g., CC-BY 4.0) should be included for each asset.
        \item For scraped data from a particular source (e.g., website), the copyright and terms of service of that source should be provided.
        \item If assets are released, the license, copyright information, and terms of use in the package should be provided. For popular datasets, \url{paperswithcode.com/datasets} has curated licenses for some datasets. Their licensing guide can help determine the license of a dataset.
        \item For existing datasets that are re-packaged, both the original license and the license of the derived asset (if it has changed) should be provided.
        \item If this information is not available online, the authors are encouraged to reach out to the asset's creators.
    \end{itemize}

\item {\bf New assets}
    \item[] Question: Are new assets introduced in the paper well documented and is the documentation provided alongside the assets?
    \item[] Answer: \answerNA{}.
    \item[] Justification: The paper does not release new assets.
    \item[] Guidelines:
    \begin{itemize}
        \item The answer NA means that the paper does not release new assets.
        \item Researchers should communicate the details of the dataset/code/model as part of their submissions via structured templates. This includes details about training, license, limitations, etc.
        \item The paper should discuss whether and how consent was obtained from people whose asset is used.
        \item At submission time, remember to anonymize your assets (if applicable). You can either create an anonymized URL or include an anonymized zip file.
    \end{itemize}

\item {\bf Crowdsourcing and research with human subjects}
    \item[] Question: For crowdsourcing experiments and research with human subjects, does the paper include the full text of instructions given to participants and screenshots, if applicable, as well as details about compensation (if any)?
    \item[] Answer: \answerNA{}.
    \item[] Justification: The paper does not involve crowdsourcing nor research with human subjects.
    \item[] Guidelines:
    \begin{itemize}
        \item The answer NA means that the paper does not involve crowdsourcing nor research with human subjects.
        \item Including this information in the supplemental material is fine, but if the main contribution of the paper involves human subjects, then as much detail as possible should be included in the main paper.
        \item According to the NeurIPS Code of Ethics, workers involved in data collection, curation, or other labor should be paid at least the minimum wage in the country of the data collector.
    \end{itemize}

\item {\bf Institutional review board (IRB) approvals or equivalent for research with human subjects}
    \item[] Question: Does the paper describe potential risks incurred by study participants, whether such risks were disclosed to the subjects, and whether Institutional Review Board (IRB) approvals (or an equivalent approval/review based on the requirements of your country or institution) were obtained?
    \item[] Answer: \answerNA{}.
    \item[] Justification: The paper does not involve crowdsourcing nor research with human subjects.
    \item[] Guidelines:
    \begin{itemize}
        \item The answer NA means that the paper does not involve crowdsourcing nor research with human subjects.
        \item Depending on the country in which research is conducted, IRB approval (or equivalent) may be required for any human subjects research. If you obtained IRB approval, you should clearly state this in the paper.
        \item We recognize that the procedures for this may vary significantly between institutions and locations, and we expect authors to adhere to the NeurIPS Code of Ethics and the guidelines for their institution.
        \item For initial submissions, do not include any information that would break anonymity (if applicable), such as the institution conducting the review.
    \end{itemize}

\item {\bf Declaration of LLM usage}
    \item[] Question: Does the paper describe the usage of LLMs if it is an important, original, or non-standard component of the core methods in this research? Note that if the LLM is used only for writing, editing, or formatting purposes and does not impact the core methodology, scientific rigorousness, or originality of the research, declaration is not required.
    \item[] Answer: \answerNA{}.
    \item[] Justification: Not involved.
    \item[] Guidelines:
    \begin{itemize}
        \item The answer NA means that the core method development in this research does not involve LLMs as any important, original, or non-standard components.
        \item Please refer to our LLM policy (\url{https://neurips.cc/Conferences/2025/LLM}) for what should or should not be described.
    \end{itemize}

\end{enumerate}

\end{document}